%% file: main.tex
\documentclass[10pt,twocolumn,letterpaper]{article}

\usepackage[left=2cm, right=2cm, top=2.5cm, bottom=2.5cm]{geometry}
\usepackage{multirow}

\usepackage{axessibility}
\usepackage{xcolor}
\usepackage{times}
\usepackage{epsfig}
\usepackage{graphicx}
\usepackage{amsmath}
\usepackage{amssymb}
\usepackage{booktabs}
\usepackage{float}
\usepackage[numbers]{natbib}
\usepackage[font=small]{caption}

\usepackage[autostyle]{csquotes} 
\usepackage[breaklinks=true,bookmarks=false]{hyperref}

\begin{document}

\title{
Image Segmentation Using Text and Image Prompts
}

\date{}

\author{
Timo Lüddecke$^{1,\S}$ and Alexander Ecker$^{1,2}$ \\
\small $^1$Institute of Computer Science and CIDAS, University of Göttingen \hspace{1cm}
\small$^2$MPI for Dynamics and Self-Organization, Göttingen
}

\maketitle

\definecolor{blue2}{rgb}{0.2,0.6,0.8}
\definecolor{cyan2}{rgb}{0.2,0.7,0.7}
\definecolor{yellow2}{rgb}{0.7,0.5,0.0}
\definecolor{blue3}{rgb}{0.3,0.5,0.7}
\definecolor{red2}{rgb}{0.6,0.1,0.2}

\def\todo#1{{ \color{blue}{TODO: #1} } }
\def\new{{ \color{red}{new} } }
\def\remark#1{{ \color{cyan}{#1} } }
\def\question#1{{ \color{orange}{Q: #1} } }

\def\conf#1{}

\newcommand{\x}{\mathbf{x}}
\newcommand{\y}{\mathbf{y}}
\newcommand{\xv}{\mathbf{x}}
\newcommand{\yv}{\mathbf{y}}
\newcommand{\tv}{\mathbf{t}}
\newcommand{\cv}{\mathbf{c}}
\newcommand{\pred}{\mathbf{p}}
\newcommand{\gt}{\mathbf{g}}
\newcommand{\miou}{\text{mIoU}}
\newcommand{\ioufg}{\text{IoU}_{\text{FG}}}
\newcommand{\ioubin}{\text{IoU}_{\text{BIN}}}

\begin{abstract}

Image segmentation is usually addressed by training a model for a fixed set of object classes. Incorporating additional classes or more complex queries later is expensive as it requires re-training the model on a dataset that encompasses these expressions.
Here we propose a system that can generate image segmentations based on arbitrary prompts at test time.
A prompt can be either a text or an image. This approach enables us to create a unified model (trained once) for three common segmentation tasks, which come with distinct challenges: referring expression segmentation, zero-shot segmentation and one-shot segmentation.
We build upon the CLIP model as a backbone which we extend with a transformer-based decoder that enables dense prediction. After training on an extended version of the PhraseCut dataset, our system generates a binary segmentation map for an image based on a free-text prompt or on an additional image expressing the query. 
We analyze different variants of the latter image-based prompts in detail.
This novel hybrid input allows for dynamic adaptation not only to the three segmentation tasks mentioned above, but to any binary segmentation task where a text or image query can be formulated.
Finally, we find our system to adapt well to generalized queries involving affordances or properties.
Code is available at \url{https://eckerlab.org/code/clipseg}.

\end{abstract}

\section{Introduction}

\let\thefootnote\relax\footnotetext{\small$^\S$timo.lueddecke@uni-goettingen.de}

The ability to generalize to unseen data is a fundamental problem relevant for a broad range of applications in artificial intelligence. For instance, it is crucial that a household robot understands the prompt of its user, which might involve an unseen object type or an uncommon expression for an object. While humans excel at this task, this form of inference is challenging for computer vision systems. 

Image segmentation requires a model to output a prediction for each pixel. Compared to whole-image classification, segmentation requires not only predicting what can be seen but also where it can be found. 
Classical semantic segmentation models are limited to segment the categories they have been trained on. 
Different approaches have emerged that extend this fairly constrained setting (see Tab.~\ref{tab:segmentation_tasks}):
\begin{itemize}
\setlength\itemsep{0em}
    \item In generalized zero-shot segmentation, seen as well as unseen categories needs to be segmented by putting unseen categories in relation to seen ones, e.g. through word embeddings \cite{word2vec} or WordNet \cite{wordnet}. 
    \item In one-shot segmentation, the desired class is provided in form of an image (and often an associated mask) in addition to the query image to be segmented.
    \item In referring expression segmentation, a model is trained on complex text queries but sees all classes during training (i.e. no generalization to unseen classes).
\end{itemize}

\begin{figure}
    \centering
    \includegraphics[width=8cm]{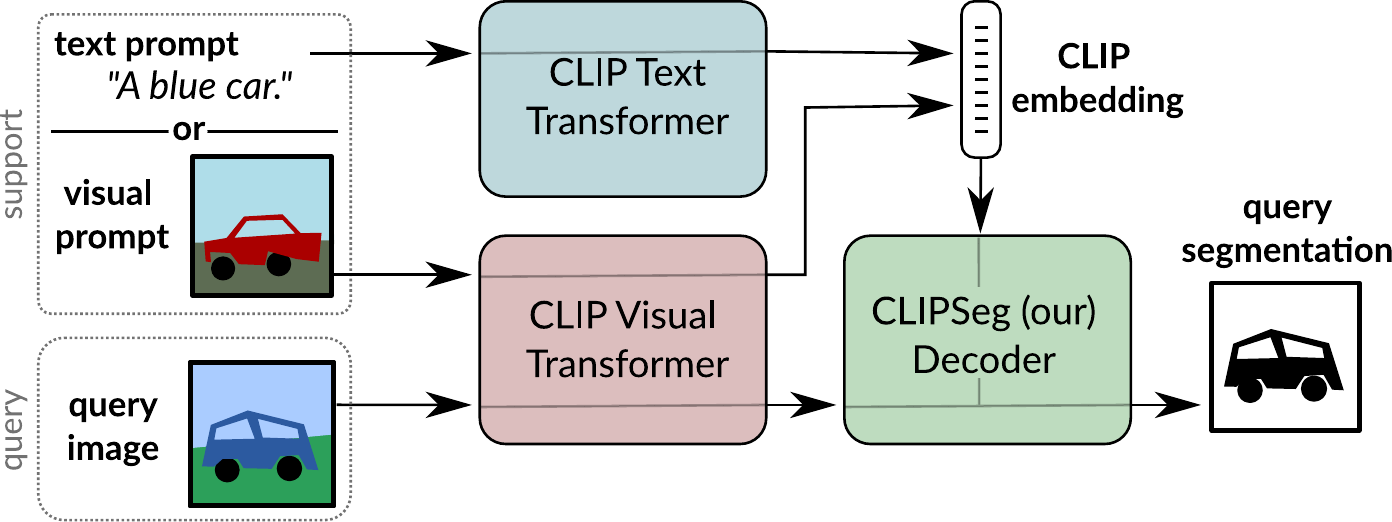}
    \caption{Our key idea is to use CLIP to build a flexible zero/one-shot segmentation system that addresses multiple tasks at once.}
    \label{fig:mini_overview}
\end{figure}

To this work, we introduce the CLIPSeg model (Fig.~\ref{fig:mini_overview}), which is capable of segmenting based on an arbitrary text query or an example image. CLIPSeg can address all three tasks named above.
This multi-modal input format goes beyond existing multi-task benchmarks such as Visual Decathlon \cite{rebuffi17learning} where input is always provided in form of images. 
To realize this system, we employ the pre-trained CLIP model as a backbone and train a thin conditional segmentation layer (decoder) on top. We use the joint text-visual embedding space of CLIP for conditioning our model, which enables us to process prompts in text form as well as images. 
Our idea is to teach the decoder to relate activations inside CLIP with an output segmentation, while permitting as little dataset bias as possible and maintaining the excellent and broad predictive capabilities of CLIP.

We employ a generic binary prediction setting, where a foreground that matches the prompt has to be differentiated from background. 
This binary setting can be adapted to multi-label predictions which is needed by Pascal zero-shot segmentation.
Although the focus of our work is on building a versatile model, we find that CLIPSeg achieves competitive performance across three low-shot segmentation tasks. Moreover, it is able to generalize to classes and expressions for which it has never seen a segmentation.

\begin{table}
    \footnotesize
    \centering
    \setlength{\tabcolsep}{1mm}
    \begin{tabular}{lcccccc}
    \toprule
        & \multirow{2}{0.9cm}{unseen classes} & \multirow{2}{1.2cm}{free form prompt} & \multirow{2}{1.1cm}{no fixed targets} & \multirow{2}{1cm}{negative samples} \\
        \\
    \midrule
        Our setting & \checkmark & \checkmark &  \checkmark & \checkmark \\
        \midrule
        Classic & - & - & - & \checkmark \\
        Referring Expression  & - & \checkmark & \checkmark & -  \\
        Zero-shot & \checkmark & -  & \checkmark & \checkmark  \\
        One-shot & \checkmark & -  & \checkmark &  - \\
    \bottomrule
    \end{tabular}
    \caption{Comparison of different segmentation tasks. 
    Negative means samples that do not contain the target (or one of the targets in multi-label segmentation). All approaches except classic segmentation adapt to new targets dynamically at inference time.
    }
    \label{tab:segmentation_tasks}
\end{table}

\paragraph{Contributions}

Our main technical contribution is the CLIPSeg model, which extends the well-known CLIP transformer for zero-shot and one-shot segmentation tasks by a proposing a lightweight transformer-based decoder. A key novelty of this model is that the segmentation target can be specified by different modalities: through text or an image. 
This allows us to train a unified model for several benchmarks. 
For text-based queries, unlike networks trained on PhraseCut, our model is able to generalize to new queries involving unseen words. For image-based queries, we explore various forms of visual prompt engineering -- analogously to text prompt engineering in language modeling.
Furthermore, we evaluate how our model generalizes to novel forms of prompts involving affordances.

\section{Related Work}

\paragraph{Foundation Models and Segmentation}

Instead of learning from scratch, modern vision systems are commonly pre-trained on a large-scale dataset (either supervised \cite{imagenet} or self-supervised \cite{chen20, chen21}) and use weight transfer. The term foundation model has been coined for very large pre-training models that are applicable to multiple downstream tasks \cite{bommasani21foundation}. One of these models is CLIP \cite{radford20}, which has demonstrated excellent performance on several image classification tasks. In contrast to previous models which rely on ResNet \cite{he16} backbones, the best-performing CLIP model uses a novel visual transformer \cite{dosovitskiy20} architecture.
Analogously to image classification, there have been efforts to make use of transformers for segmentation: 
\mbox{TransUNet} \cite{chen21transunet} and SETR \cite{zheng21setr} employ a hybrid architecture which combine a visual transformer for encoding with a CNN-based decoder.
Segformer \cite{xie21segformer} combines a transformer encoder with an MLP-based decoder.
The Segmentor model \cite{strudel21segmenter} pursues a purely transformer-based approach. To generate a segmentation, either a projection of the patch embeddings or mask transformer are proposed. 
Our CLIPSeg model extends CLIP with a transformer-based decoder, i.e. we do not rely on convolutional layers.

\paragraph{Referring Expression Segmentation}

In referring expression segmentation a target is specified in a natural language phrase. The goal is to segment all pixels that match this phrase. Early approaches used recurrent networks in combination with CNNs to address this problem \cite{hu2016seg_nl, liu17refseg, shi18, li18referring}.
The CMSA module, which is central to the approach of \citet{ye19}, models long-term dependencies between text and image using attention.
The more recent HULANet method \cite{wu20phrasecut} consists of Mask-RCNN backbone and specific modules processing categories, attributes and relations, which are merged to generate a segmentation mask.
MDETR \cite{kamath21} is an adaptation of the detection method DETR \cite{carion20end} to natural language phrase input. It consists of a CNN which extracts features and a transformer which predicts bounding boxes for a set of query prompts. 
Note that referring expression segmentation does not require generalization to unseen object categories or understanding of visual support images.
Several benchmarks \cite{yu16modeling, mao16generation, wu20phrasecut} were proposed to track progress in referring expression segmentation. We opt for the PhraseCut dataset \cite{wu20phrasecut} which is substantially larger in terms of images and classes than other datasets. It contains structured text queries involving objects, attributes and relationships. A query can match multiple object instances.

\paragraph{Zero-Shot Segmentation}

In zero-shot segmentation the goal is to segment objects of categories that have not been seen during training. Normally, multiple classes need to be segmented in an image at the same time. In the generalized setting, both seen and unseen categories may occur. 
A key problem in zero-shot segmentation addressed by several methods is the bias which favors seen classes.
\citet{bucher19} train a DeepLabV3-based network to synthesize artificial, pixel-wise features for unseen classes based on word2vec label embeddings. These features are used to learn a classifier. 
Follow-up work explicitly models the relation between seen and unseen classes  \cite{li20}.
Others add semantic class information into dense prediction models \cite{xian19}.
More recent approaches use a joint space for image features and class prototypes \cite{baek21}, employ a probabilistic formulation to account for uncertainty \cite{hu20} or model the detection of unseen objects explicitly \cite{zhang21}.

\paragraph{One-Shot Semantic Segmentation}
In one-shot semantic segmentation, the model is provided at test time with a single example of a certain class, usually as an image with a corresponding mask. 
One-shot semantic segmentation is a comparably new task, with the pioneering work being published in 2017 by Shaban et~al. \cite{shaban17}, which introduced the Pascal-5i dataset based on Pascal images and labels.
Their simple model extracts VGG16-features \cite{simonyan14} from a masked support image to generate regression parameters that are applied per-location on the output of a FCN \cite{long15} to yield a segmentation. 
Later works introduce more complex mechanisms to handle one-shot segmentation:
The pyramid graph network (PGNet) \cite{zhang19d} generates a set of differently-shaped feature maps obtained through adaptive pooling and processes them by individual graph attention units and passed through an atrous spatial pyramid pooling (ASPP) block \cite{chen18a}.
The CANet network \cite{zhang19a} first extracts images using a shared encoder. Then predictions are iteratively refined through a sequence of convolutions and ASPP blocks.
Several approaches focus on the modeling of prototypes \cite{wang19a, yang20, liu20}.
PFENet \cite{tian20a} uses a prior computed on high-level CNN-features to provide an auxiliary segmentation that helps further processing.
A weakly-supervised variant as introduced by Rakelly et~al. \cite{rakelly18fewshot} requires only sparse annotations in form of a set of points.
In one-shot instance segmentation \cite{michaelis18}, instead of a binary match/non-match prediction, individual object instances are segmented.

\paragraph{CLIP Extensions}
Despite CLIP \cite{radford20} being fairly new, multiple derivative works across different sub-fields have emerged.
CLIP was combined with a GAN to modify images based on a text prompt \cite{patashnik21styleclip} and
in robotics to generalize to unseen objects in manipulations tasks \cite{shridhar21cliport}.
Other work focused on understanding CLIP in more detail. In the original CLIP paper \cite{radford20}, it was found that the design of prompts matters for downstream tasks, i.e. instead of using an object name alone as a prompt, adding the prefix ``a photo of" increases performance. \citet{zhou21coop} propose context optimization (CoOp) which automatically learns tokens that perform well for given downstream tasks.
Other approaches rely on CLIP for open-set object detection \cite{gu21zero, esmaeilpour21zero}.

\begin{figure*}
\centering
 \includegraphics[width=0.8\textwidth]{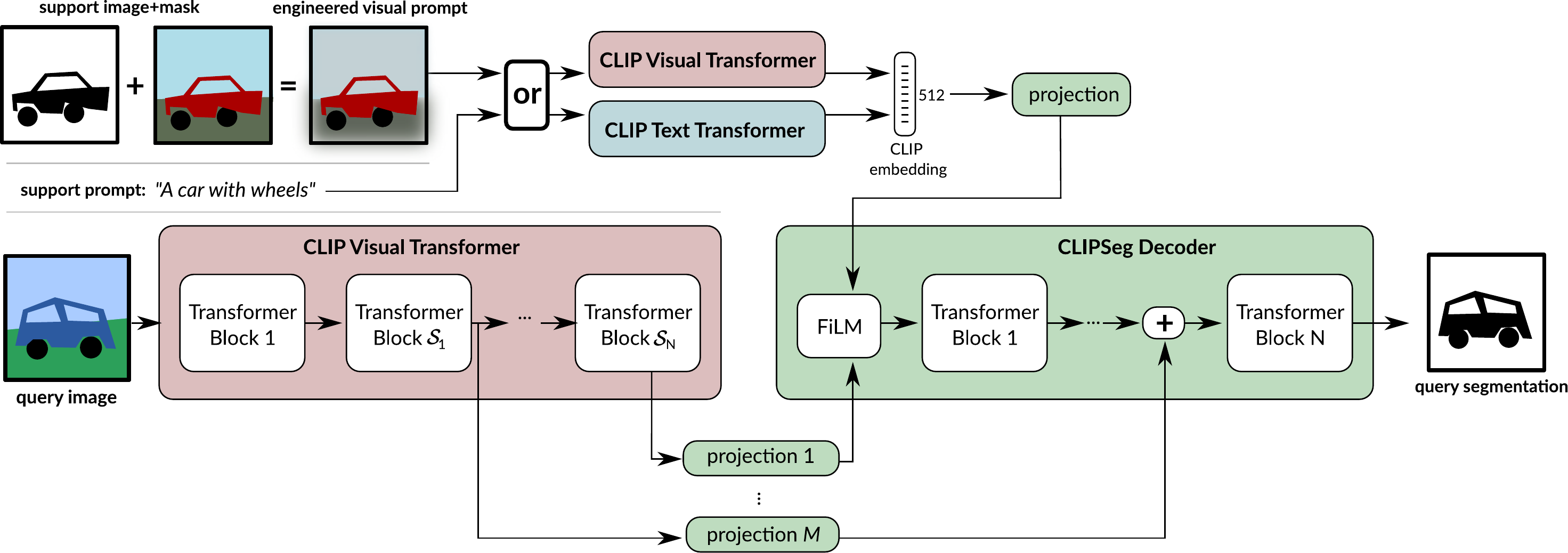}
\caption{Architecture of CLIPSeg: We extend a frozen CLIP model (red and blue) with a transformer that segments the query image based on either a support image or a support prompt. $N$ CLIP activations are extracted after blocks defined by $\mathcal{S}$. The segmentation transformer and the projections (both green) are trained on PhraseCut or PhraseCut+.}\label{fig:clipseg}
\end{figure*}

\begin{figure}
\centering
\includegraphics[width=0.4\textwidth]{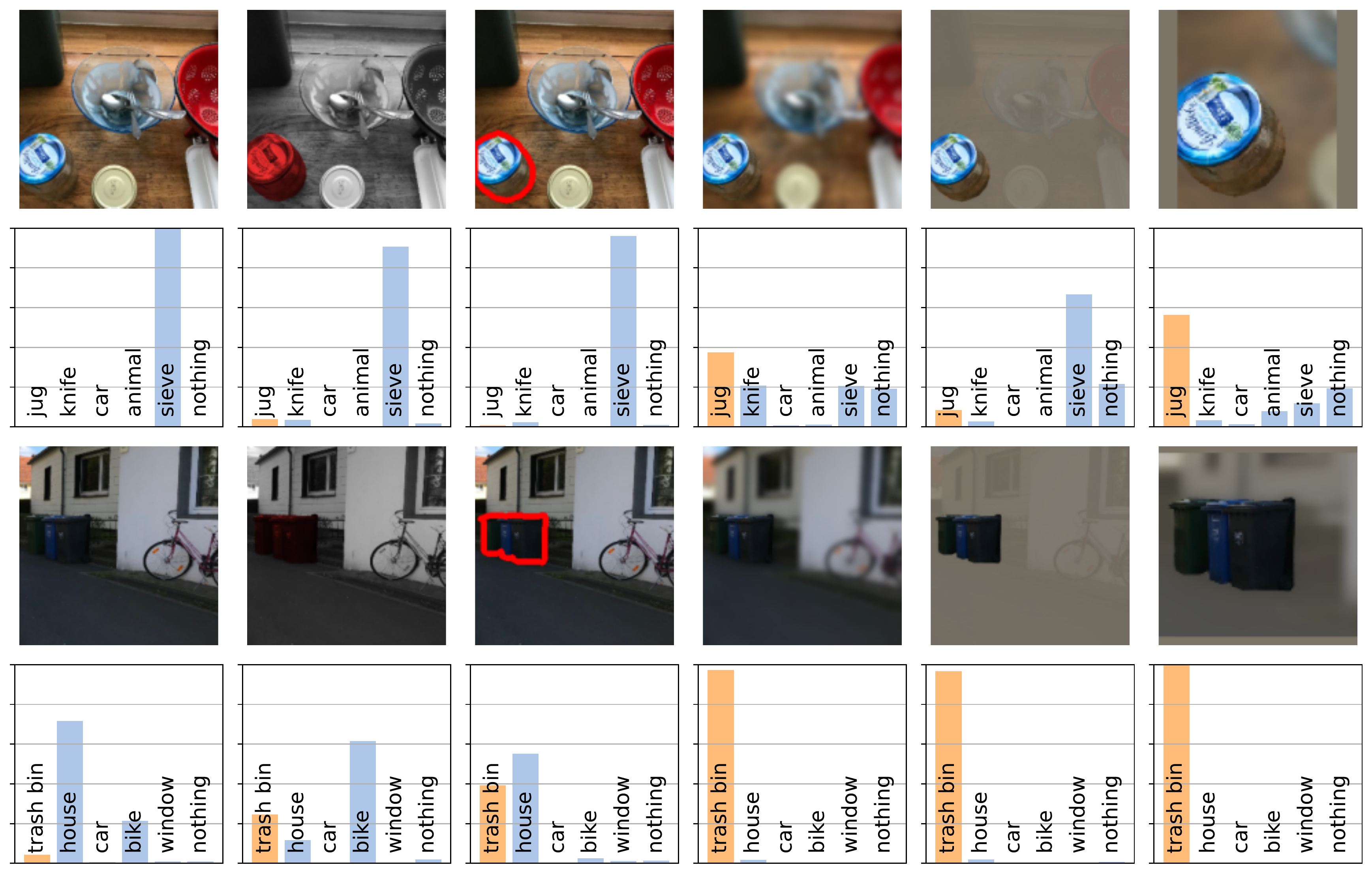}
\caption{Different forms of combining an image with the associated object mask to build a visual prompt have a strong effect on CLIP predictions (bar charts). We use the difference in the probability of the target object (orange) in the original image (left column) and the masking methods for our systematic analysis.}
\label{fig:prompt_engineering}
\end{figure}

\section{CLIPSeg Method}

We use the visual transformer-based (ViT-B/16) CLIP \cite{radford20} model as a backbone (Fig.~\ref{fig:clipseg}) and extend it with a small, parameter-efficient transformer decoder. The decoder is trained on custom datasets to carry out segmentation, while the CLIP encoder remains frozen.
A key challenge is to avoid imposing strong biases on predictions during segmentation training and maintaining the versatility of CLIP. We do not use the larger ViT-L/14@336px CLIP variant as its weights were not publicly released as of writing this work.

\def\v{\mathbf{v}}
\def\t{\mathbf{t}}
\def\y{\mathbf{y}}
\def\q{\mathbf{q}}
\def\s{\mathbf{s}}

\paragraph{Decoder Architecture}

Considering these demands, we propose CLIPSeg: A simple, purely-transformer based decoder, which has U-Net-inspired skip connections to the CLIP encoder that allow the decoder to be compact (in terms of parameters). 
While the query image ($\mathbb{R}^{W \times H \times 3}$) is passed through the CLIP visual transformer, activations at certain layers $\mathcal{S}$ are read out and projected to the token embedding size $D$ of our decoder. 
Then, these extracted activations (including CLS token) are added to the internal activations of our decoder before each transformer block. The decoder has as many transformer blocks as extracted CLIP activations (in our case 3).
The decoder generates the binary segmentation by applying a linear projection on the tokens of its transformer (last layer) $\mathbb{R}^{(1 + \frac{W}{P} \times \frac{H}{P}) \times D} \mapsto  \mathbb{R}^{W \times H}$, where $P$ is the token patch size of CLIP.
In order to inform the decoder about the segmentation target, we modulate the decoder's input activation by a conditional vector using FiLM \cite{dumoulin18}. 
This conditional vector can be obtained in two ways: (1) Using the CLIP text-transformer embedding of a text query and (2) using the CLIP visual transformer on a feature engineered prompt image. 
CLIP itself is not trained, but only used as a frozen feature extractor. Due to the compact decoder, CLIPSeg has only 1,122,305 trainable parameters for $D=64$.

The original CLIP is constrained to a fixed image size due to the learned positional embedding. We enable different image sizes (including larger ones) by interpolating the positional embeddings. To validate the viability of this approach, we compare prediction quality for different image sizes and find that for ViT-B/16 performance only decreases for images larger than 350 pixels (see supplementary for details).
In our experiments we use CLIP ViT-B/16 with a patch size $P$ of 16 and use a projection dimension of $D = 64$ if not indicated otherwise. We extract CLIP activations at layers $\mathcal{S} = [3,7,9]$, consequently our decoder has only three layers.

\paragraph{Image-Text Interpolation}
\label{sec:text_image_interpolation}
Our model receives information about the segmentation target (``what to segment?'') through a conditional vector. This can be provided either by text or an image (through visual prompt engineering). Since CLIP uses a shared embedding space for images and text captions, we can interpolate between both in the embedding space and condition on the interpolated vector.
Formally, let $\s_i$ be the embedding of the support image and $\t_i$ the text embedding of a sample $i$, we obtain a conditional vector $\x_i$ by a linear interpolation $\x_i= a \s_i + (1-a) \x_i$, where $a$ is sampled uniformly from $[0, 1]$. 
We use this randomized interpolation as a data augmentation strategy during training.

\subsection{PhraseCut + Visual prompts (PC+)}

We use the PhraseCut dataset \cite{wu20phrasecut}, which encompasses over 340,000 phrases with corresponding image segmentations.
Originally, this dataset does not contain visual support but only phrases and for every phrase a corresponding object exists. We extend this dataset in two ways: visual support samples and negative samples. 
To add visual support images for a prompt $p$, we randomly draw from the set of all samples $\mathcal{S}_p$, which share the prompt $p$. In case the prompt is unique ($|\mathcal{S}_p| = 1$), we rely only on the text prompt. 
Additionally, we introduce negative samples to the dataset, i.e. samples in which no object matches the prompt. To this end, the sample's phrase is replaced by a different phrase with a probability $q_{neg}$.
Phrases are augmented randomly using a set of fixed prefixes (as suggested by the CLIP authors). On the images we apply random cropping under consideration of object locations, making sure the object remains at least partially visible.
In the remainder of this paper, we call this extended dataset PhraseCut+ (abbreviated by PC+). In contrast to the original PhraseCut dataset, which uses only text to specify the target, PC+ supports training using image-text interpolation. This way, we can train a joint model that operates on text and visual input.

\section{Visual Prompt Engineering}

\label{sec:prompt_engineering}

\begin{table*}
    \centering
    \footnotesize
    \begin{tabular}{lr}
        \toprule
         CLIP modification \& extras & $\Delta \text{P}(\text{object})$ \\
         \midrule
CLIP masking CLS in layer 11 & 1.34 \\
CLIP masking CLS in all layers & 1.71 \\
CLIP masking all in all layers & -14.44 \\
dye object red in grays. image & 1.21 \\
add red object outline & 2.29 \\
         \bottomrule
    \end{tabular}
    \hspace{0.5cm}
    \begin{tabular}{lr}
        \toprule
        background modific. & $\Delta \text{P}(\text{object})$ \\
         \midrule
BG intensity 50\% & 3.08 \\
BG intensity 10\% & 13.85 \\
BG intensity 0\% & 23.40 \\
BG blur & 13.15 \\
+ intensity 10\% & 21.73 \\
         \bottomrule
    \end{tabular} 
    \hspace{0.5cm}
    \begin{tabular}{lr}
        \toprule
         cropping \& combinations & $\Delta \text{P}(\text{object})$ \\
         \midrule
crop large context & 6.27 \\
crop & 13.60 \\
crop \& BG blur & 15.34 \\
crop \& BG intensity 10\% & 21.73 \\
+ BG blur & \textbf{23.50} \\
         \bottomrule
    \end{tabular}        
    \caption{Visual prompt engineering: Average improvement of object probability for different forms of combining image and mask over 1,600 samples. Cropping means cutting the image according to the regions specified by the mask, ``BG'' means background.}
    \label{tab:prompt_engineering_short}
\end{table*}

In conventional, CNN-based one-shot semantic segmentation, masked pooling \cite{shaban17} has emerged as a standard technique to compute a prototype vector for conditioning. The provided support mask is downsampled and multiplied with a late feature map from the CNN along the spatial dimensions and then pooled along the spatial dimensions. This way, only features that pertain to the support object are considered in the prototype vector.
This method cannot be applied directly to transformer-based architectures, as semantic information is also accumulated in the CLS token throughout the hierarchy and not only in the feature maps. 
Circumventing the CLS token and deriving the conditional vector directly from masked pooling of the feature maps is not possible either, since it would break the compatibility between text embeddings and visual embeddings of CLIP.

To learn more about how target information can be incorporated into CLIP, we compare several variants in a simple experiment without segmentation and its confounding effects.
We consider the cosine distance (alignment) between visual and text-based embedding and use the original CLIP weights without any additional training.

Specifically, we use CLIP to compute the text embeddings $\mathbf{t}_i$ which correspond to object names in the image. We then compare those to (1) the visual embedding of the original image without modifications, $\mathbf{s}_{\text{o}}$ and (2) the visual embedding $\mathbf{s}_{\text{h}}$ highlighting the target object using a modified RGB image or attention mask (both techniques are described in detail below).
By softmax-normalizing the vector of alignments $[\mathbf{s}_{\text{h}} \mathbf{t}_0, \mathbf{s}_{\text{h}} \mathbf{t}_1, \dots]$ for different highlighting techniques and images, we obtain the distributions shown in Fig.~\ref{fig:prompt_engineering}.
For quantitative scores, we consider only the target object name embedding $\mathbf{t}_0$, which we expect to have a stronger alignment with the highlighted image embedding $\mathbf{s}_{\text{h}}$ than with the original image embedding $\mathbf{s_0}$ (Fig.~\ref{fig:prompt_engineering}). This means, if a highlighting technique improves the alignment, the increase in object probability $\Delta \text{P}(\text{object}) = \mathbf{s}_{\text{h}} \mathbf{t}_0  - \mathbf{s}_{o} \mathbf{t}_0 $ should be large.
We base this analysis on the LVIS dataset \cite{gupta19lvis} since its images contain multiple objects and a rich set of categories. We sample 1,600 images and mask one target object out of all objects present in this image.

\paragraph{CLIP-Based Masking}
The straightforward equivalent to masked pooling in a visual transformer is to apply the mask on the tokens. Normally, a visual transformer consists of a fixed set of tokens which can interact at every layer through multi-head attention: A CLS token used for read-out and image-region-related tokens which were originally obtained from image patches. 
Now, the mask can be incorporated by constraining the interaction at one (e.g. the last layer 11) or more transformer layers to within-mask patch tokens as well as the CLS token only.
Our evaluation (Tab.~\ref{tab:prompt_engineering_short}, left) suggests that this form of introducing the mask does not work well. By constraining the interactions with the CLS token (Tab.~\ref{tab:prompt_engineering_short}, left, top two rows) only a small improvement is achieved (in last layer or in all layers) while constraining all interactions decreases performance dramatically. From this we conclude that more complex strategies are necessary to combine image and mask internally.

\paragraph{Visual Prompt Engineering}
Instead of applying the mask inside the model, we can also combine mask and image to a new image, which can then processed by the visual transformer.
Analogous to prompt engineering in NLP (e.g. in GPT-3 \cite{brown20}), we call this procedure visual prompt engineering. Since this form of prompt design is novel and strategies which perform best in this context are unknown, we conduct an extensive evaluation of different variants of designing visual prompts (Tab.~\ref{tab:prompt_engineering_short}).
We find that the exact form of how the mask and image are combined matters a lot. Generally, we identify three image operations that improve the alignment between the object text prompts and the images: decreasing the background brightness, blurring the background (using a Gaussian filter) and cropping to the object. The combination of all three performs best (Tab.~\ref{tab:prompt_engineering_short}, last row). We will use this variant in the remainder.

\section{Experiments}

We first evaluate our model on three established segmentation benchmarks before demonstrating the main contribution of our work: flexible few-shot segmentation that can be based on either text or image prompts.

\paragraph{Metrics}

Compared to approaches in zero-shot and one-shot segmentation (e.g. \cite{bucher19, li20}), the vocabulary we use is open, i.e. the set of classes or expressions is not fixed. Therefore, throughout the experiments, our models are trained to generate binary predictions that indicate where objects matching the query are located. If necessary, this binary setting can be transformed into a multi-label setting (as we do in Section~\ref{sec:zeroshot}).

In segmentation, intersection over union (IoU, also Jaccard score) is a common metric to compare predictions with ground truth. 
Due to the diversity of the tasks, we employ different forms of IoU: Foreground IoU ($\ioufg$) which computes IoU on foreground pixels only, mean IoU, which computes the average over foreground IoUs of different classes and binary IoU ($\ioubin$) which averages over foreground IoU and background IoU.
In binary segmentation, IoU requires a threshold $t$ to be specified. While most of the time the natural choice of 0.5 is used, the optimal values can strongly deviate from 0.5 if the probability that an object matching the query differs between training and inference (the a-priori probability of a query matching one or more objects in the scene depends highly on context and dataset). 
Therefore, we report performance of one-shot segmentation using thresholds $t$ optimized per task and model. Additionally, we adopt the average precision metric (AP) in all our experiments.
Average precision measures the area under the recall-precision curve. It measures how well the system can discriminate matches from non-matches, independent of the choice of threshold. 

\paragraph{Models and Baselines}

In our experiments we differentiate two variants of CLIPSeg: One trained on the original PhraseCut dataset (PC) and one trained on the extended version of PhraseCut which uses 20\% negative samples, contains visual samples (PC+) and uses image-text interpolation (Sec.~\ref{sec:text_image_interpolation}). The robust latter version we call the universal model. To put the performance of our models into perspective, we provide two baselines:
\begin{itemize}
\setlength\itemsep{0em}
    \item \emph{CLIP-Deconv} encompasses CLIP but uses a very basic decoder, consisting only of the basic parts: FiLM conditioning \cite{dumoulin18}, a linear projection and a deconvolution. This helps us to estimate to which degree CLIP-alone is responsible for the results.
    \item \emph{ViTSeg} shares the architecture of CLIPSeg, but uses an ImageNet-trained visual transformer as a backbone \cite{timm}. For encoding text, we use the same text transformer of CLIP. This way we learn to which degree the specific CLIP weights are crucial for good performance.
\end{itemize}
We rely on PyTorch \cite{pytorch} for training and use an image size of 352 $\times$ 352 pixels throughout our experiments (for details see appendix).

\subsection{Referring Expression Segmentation}
We evaluate referring expression segmentation performance (Tab.~\ref{tab:ref_seg}) on the original PhraseCut dataset and compare to scores reported by \citet{wu20phrasecut} as well as the concurrently developed transformer-based MDETR method \cite{kamath21}. 
For this experiment we trained a version of CLIPSeg on the original PhraseCut dataset (CLIPSeg [PC]) using only text labels in addition to the universal variant which also includes visual samples (CLIPSeg [PC+]).

Our approaches outperform the two-stage HULANet approach by Wu et al. \cite{wu20phrasecut}. Especially, a high capacity decoder ($D=128$) seems to be beneficial for PhraseCut. However, the performance worse than MDETR \cite{kamath21}, which operates at full image resolution and received two rounds of fine-tuning on PhraseCut. 
Notably, the ViTSeg baseline performs generally worse than CLIPSeg, which shows that CLIP pre-training is helpful.

\begin{table}[]
    \centering
    \footnotesize
    \begin{tabular}{llllllll}
        \toprule
        & $t$ &  $\miou$ & $\ioufg$ & AP \\
        \midrule
        CLIPSeg (PC+)    &  0.3 &  43.4 &  54.7 &  76.7 \\
 CLIPSeg (PC, $D=128$) &  0.3 &  48.2 &  \textbf{56.5} &  \textbf{78.2} \\
          CLIPSeg (PC)   &  0.3 &  46.1 &  56.2 &  \textbf{78.2} \\
           CLIP-Deconv   &  0.3 &  37.7 &  49.5 &  71.2 \\
         ViTSeg (PC+)    &  0.1 &  28.4 &  35.4 &  58.3 \\
           ViTSeg (PC)   &  0.3 &  38.9 &  51.2 &  74.4 \\
        \midrule
        MDETR \cite{kamath21} & \conf{ICCV21} & \textbf{53.7} & -  & - \\
       HulaNet \cite{wu20phrasecut} & \conf{CVPR20}  & 41.3 & 50.8   & -\\
       Mask-RCNN top \cite{wu20phrasecut} & \conf{CVPR20} & 39.4 & 47.4  & - \\
       RMI \cite{wu20phrasecut} & \conf{CVPR20} & 21.1 & 42.5  & - \\
        \bottomrule
    \end{tabular}
    \caption{Referring Expression Segmentation performance on PhraseCut ($t$ refers to the binary threshold).}
    \label{tab:ref_seg}
\end{table}

\subsection{Generalized Zero-Shot Segmentation}
\label{sec:zeroshot}

In generalized zero-shot segmentation, test images contain categories that have never been seen before in addition to known categories. 
We evaluate the model's zero-shot segmentation performance using the established Pascal-VOC benchmark (Tab.~\ref{tab:zero_shot_performance}). It contains five splits involving 2 to 10 unseen classes (we report only 4 and 10 unseen classes). The latter is the most challenging setting as the set of unseen classes is large.
Since our model was trained on foreground/background segmentation we cannot directly use it in a multi-label setting. Therefore, we employ a simple adaptation: Our model predicts a binary map independently for each of the 20 Pascal classes. Across all 20 predictions we determine the class with the highest probability for each pixel. 

We train on PhraseCut+ but remove the unseen Pascal classes from the dataset. This is carried out by assigning the Pascal classes to WordNet synsets \cite{wordnet} and generating a set of invalid words by traversing hyponyms (e.g. different dog breeds for dog). Prompts that contain such a word are removed from the dataset.

The idea of conducting this experiment is to provide a reference for the zero-shot performance of our universal model. It should not considered as competing in this benchmark as we use a different training (CLIP pre-training, binary segmentation on PhraseCut).
The results (Tab.~\ref{tab:zero_shot_performance}) indicate a major gap between seen and unseen classes in models trained on Pascal-VOC, while our models tend to be more balanced. This is due to other models being trained exclusively on the 10 or 16 seen Pascal classes in contrast to CLIPSeg which can differentiate many more classes (or phrases). 
In fact, our model performs better on unseen classes than on seen ones. This difference is likely because the seen classes are generally harder to segment: For the unseen-4 setting, the unseen classes are ``airplane'', ``cow'', ``motorbike'' and ``sofa''. All of them are large and comparatively distinct objects.

\begin{table}[]
    \centering
    \footnotesize
    \setlength{\tabcolsep}{1.5mm}
    \begin{tabular}{ll|rr|rrll}
         \toprule
         && \multicolumn{2}{c|}{unseen-10} & \multicolumn{2}{c}{unseen-4} \\
        & pre-train. & mIoU$_S$ & mIoU$_U$ & mIoU$_S$ & mIoU$_U$  \\
        \midrule
CLIPSeg (PC+) & CLIP &  \textbf{35.7} & \textbf{43.1} & 20.8 & \textbf{47.3} \\
CLIP-Deconv (PC+) & CLIP &  25.1 & 36.7 & \textbf{25.9} & 41.9 \\
ViTSeg (PC+) & IN &  4.2 & 19.0 & 6.0 & 24.8 \\
        \midrule
        \midrule
        SPNet \cite{xian19} & IN & 59.0 & 18.1 & 67.3 & 21.8 \\
        ZS3Net \cite{bucher19} & IN-seen & 33.9 & 18.1 & 66.4 & 23.2 \\
        CSRL \cite{li20consistent}  & IN-seen  & 59.2 & 21.0 & 69.8 & 31.7  \\
        CaGNet \cite{gu2020context} & IN & - & - & 69.5 & 40.2 \\
        OSR \cite{zhang21} & IN-seen &  \textbf{72.1} & \textbf{33.9} & \textbf{75.0} & \textbf{44.1} \\
        JoEm \cite{baek21} & IN-seen & 63.4 & 22.5 & 67.0 & 33.4  \\
        \bottomrule
    \end{tabular}%
    \caption{Zero-shot segmentation performance on Pascal-VOC with 10 unseen classes. mIoU$_S$ and mIoU$_U$ indicate performance on seen and unseen classes, respectively. Our model is trained on PhraseCut with the Pascal classes being removed but uses a pre-trained CLIP backbone. IN-seen indicates ImageNet pre-training with unseen classes being removed.}
    \label{tab:zero_shot_performance}
\end{table}

\subsection{One-Shot Semantic Segmentation}

In one-shot semantic segmentation, a single example image along with a mask is presented to the network. Regions that pertain to the class highlighted in the example image must be found in a query image. Compared to previous tasks, we cannot rely on a text label but must understand the provided support image.
Above (Sec.~\ref{sec:prompt_engineering}) we identified the best method for visual prompt design, which we use here: cropping out the target object while blurring and darkening the background. 
To remove classes that overlap with the respective subset of Pascal during training, we use the same method as in the previous section (Sec.~\ref{sec:zeroshot}).
Other than in zero-shot segmentation, in one-shot segmentation, ImageNet pre-trained backbones are common \cite{tian20a, wang19a}. PFENet particularly leverages pre-training by using high-level feature similarity as a prior. Similarly, HSNet \cite{min21hsnet} processes correlated activations of query and support image using 4D-convolutions at multiple levels.

On Pascal-5i we find our universal model CLIPSeg (PC+) to achieve competitive performance (Tab.~\ref{tab:pas5i_oneshot}) among state-of-the-art methods, with only the very recent HSNet performing better.
The results on COCO-20i (Tab.~\ref{tab:coco20i_oneshot}) show that CLIPSeg also works well when trained on other datasets than PhraseCut(+). Again HSNet performs better. To put this in perspective, it should be considered that HSNet (and PFENet) are explicitly designed for one-shot segmentation, rely on pre-trained CNN activations and cannot handle text by default:
\citet{tian20a} extended PFENet to zero-shot segmentation (but used the one-shot protocol) by replacing the visual sample with word vectors \cite{word2vec, mikolov18} of text labels. In that case, CLIPSeg outperforms their scores by a large margin (Tab.~\ref{tab:pas5i_zeroshot}).

\begin{table}[t]
    \centering
    \footnotesize
    \begin{tabular}{llrrrrr}
         \toprule
        & $t$ & vis. backb. & $\miou$ & $\ioubin$ &  AP \\
        \midrule
CLIPSeg (PC+) & 0.3 & ViT (CLIP) & \textbf{59.5} & \textbf{75.0} & \textbf{82.3} \\
CLIPSeg (PC)  & 0.3 & ViT (CLIP) & 52.3 & 69.5 & 72.4 \\
CLIP-Deconv (PC+) & 0.2 & ViT (CLIP) & 48.0 & 65.8 & 68.0 \\
ViTSeg (PC+) & 0.2 & ViT (IN) & 39.0 & 59.0 & 62.4 \\
        \midrule
        PPNet  \cite{liu20} & & RN50 & 52.8  & 69.2  & - \\
        RePRI \cite{boudiaf20} & & RN50 & 59.7 & - & -  \\
        PFENet \cite{tian20a} & & RN50 & 60.2  & 73.3 & -  \\
        HSNet \cite{min21hsnet} & & RN50 & \textbf{64.0} & \textbf{76.7} & -\\
        \midrule
        PPNet  \cite{liu20} & & RN101 & 55.2  & 70.9 & -  \\
        RePRI \cite{boudiaf20} & & RN101  & 59.4  & - & - \\
        PFENet \cite{tian20a} & & RN101 & 59.6 & 72.9 & - \\
        HSNet \cite{min21hsnet} & & RN101 & \textbf{66.2} & \textbf{77.6} & - \\
        \bottomrule
         \end{tabular} %
    \caption{One-shot performance on Pascal-5i (CLIPSeg and ViTSeg trained on PhraseCut+).}
    \label{tab:pas5i_oneshot}
\end{table}

\begin{table}[t]
    \centering
    \footnotesize
    \setlength{\tabcolsep}{1.7mm}
    \begin{tabular}{llrrrr}
        \midrule
        & $t$ & vis. backb. & $\miou$ & $\ioubin$ &  AP  \\
        \midrule
CLIPSeg (COCO) & 0.1 & ViT (CLIP) &  33.2 & 58.4 & 40.5 \\
CLIPSeg (COCO+N)  & 0.1 & ViT (CLIP) &  \textbf{33.3} & \textbf{59.1} & \textbf{41.7} \\
CLIP-Deconv (COCO+N)  & 0.1 & ViT (CLIP) &  29.8 & 56.8 & 40.8 \\
ViTSeg (COCO)  & 0.1 & ViT (IN) &  14.4 & 46.1 & 15.7 \\
        \midrule
        PPNet \cite{liu20} & & RN50 & 29.0  & - & -  \\
        RePRI \cite{boudiaf20} & & RN50  & 34.0  & - & - \\
        PFENet \cite{tian20a} & & RN50 & 35.8 & - & - \\
        HSNet \cite{min21hsnet} & & RN50 & \textbf{39.2} & \textbf{68.2} & - \\
        \midrule
        HSNet \cite{min21hsnet} & & RN101 & \textbf{41.2} & \textbf{69.1} & - \\
        \bottomrule
    \end{tabular} %
    \caption{One-shot performance on COCO-20i (CLIPSeg trained on PhraseCut), +N indicates 10\% negative samples.}
    \label{tab:coco20i_oneshot}
\end{table}

\begin{table}[t]
    \centering
    \footnotesize
    \setlength{\tabcolsep}{1mm}
    \begin{tabular}{llrrrrr}
         \toprule 
        \textbf{Pascal-5i} & $t$ & vis. backb. & $\miou$ & $\ioubin$ &  AP \\
        \midrule
CLIPSeg (PC+) & 0.3 & ViT (CLIP) & \textbf{72.4} & \textbf{83.1} & \textbf{93.5} \\
CLIPSeg (PC) & 0.3 & ViT (CLIP) & 70.3 & 81.6 & 84.8 \\
CLIP-Deconv (PC+) & 0.3 & ViT (CLIP) & 63.2 & 77.3 & 85.3 \\
ViTSeg (PC+) & 0.2 & ViT (IN) & 39.0 & 59.0 & 62.4 \\
        \midrule
        LSeg  \cite{li2022languagedriven} & & ViT (CLIP) & 52.3 & 67.0 & -  \\
        PFENet \cite{tian20a} & & VGG16 & 54.2 & - & - \\
        \bottomrule
    \end{tabular}%
    \caption{Zero-shot performance on Pascal-5i. The scores were obtained by following the evaluation protocol of one-shot segmentation but using text input.}
    \label{tab:pas5i_zeroshot}
\end{table}

\subsection{One Model For All: Generalized Prompts}

\label{sec:generalize}

We have shown that CLIPSeg performs well on a variety of academic segmentation benchmarks. Next, we evaluate its performance ``in the wild" in unseen situations.

\paragraph{Qualitative Results}

\begin{figure*}[tb]
    \centering
    \includegraphics[width=0.98\textwidth]{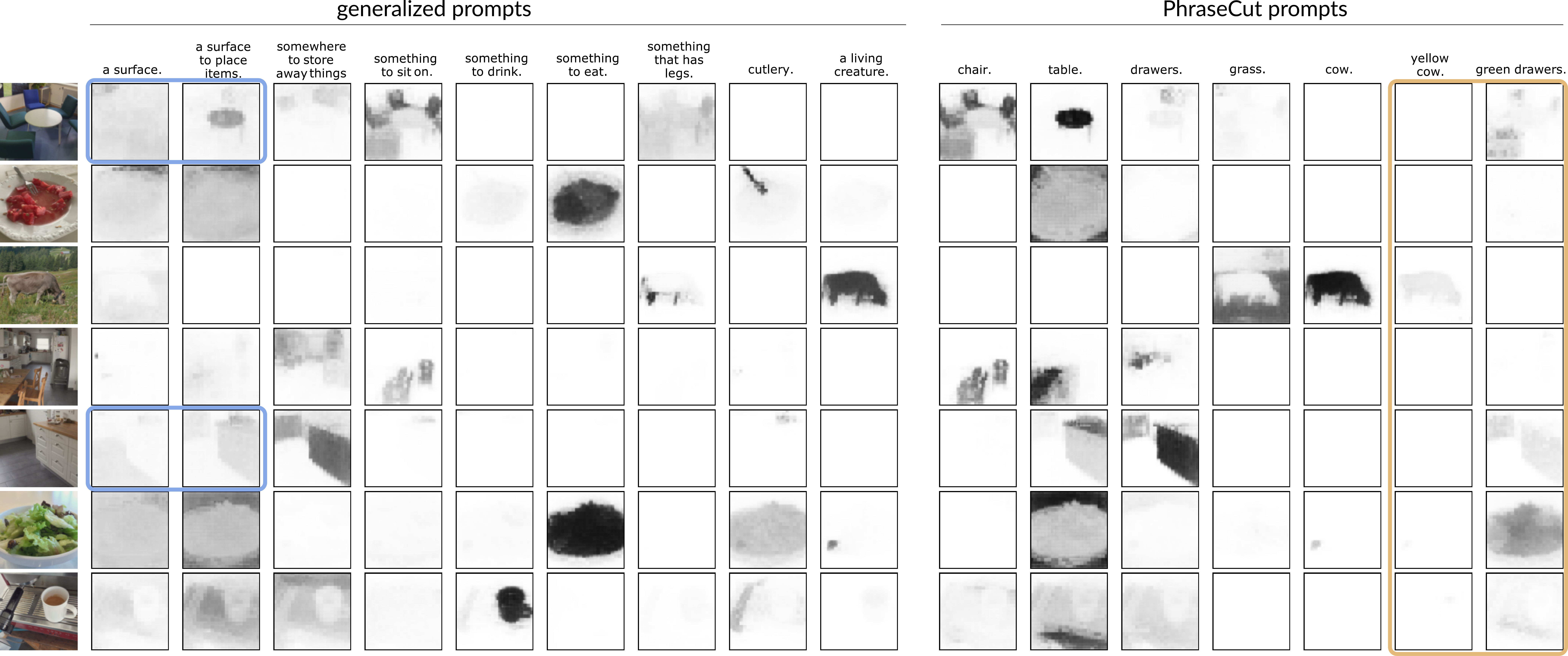}
    \caption{Qualitative predictions of CLIPSeg (PC+) for various prompts, darkness indicates prediction strength. The generalized prompts (left) deviate from the PhraseCut prompts as they involve action-related properties or new object names.}
    \label{fig:text_qualitative}
\end{figure*}

In Fig.~\ref{fig:text_qualitative} we show qualitative results divided into two groups: (1, left) Affordance-like \cite{gibson66, gibson79} (``generalized'')  prompts that are different from the descriptive prompts of PhraseCut and (2, right) prompts that were taken from the PhraseCut test set. For the latter we add challenging extra prompts involving an existing object but the wrong color (indicated in orange). 
Generalized prompts, which deviate from the PhraseCut training set by referring to actions (```something to ...'') or rare object classes (```cutlery'') work surprisingly well given that the model was not trained on such cases. It has learned an intuition of stuff that can be stored away in cupboards, where sitting is possible and what ``living creature'' means. Rarely, false positives are generated (the bug in the salad is not a cow).
Details in the prompt are reflected by the segmentation (blue boxes) and information about the color influences predicted object probabilities strongly (orange box).

\paragraph{Systematic Analysis}
To quantitatively assess the performance for generalized queries, we construct subsets of the LVIS test datasets containing only images of classes that correspond to affordances or attributes.
Then we ask our model to segment with these affordances or attributes as prompts. For instance, we compute the foreground intersection of union between armchair, sofa and loveseat objects when ``sit on'' is used as prompt.
A complete list of which affordances or attributes are mapped onto which objects can be found in the appendix.
We find (Tab.~\ref{tab:generalize}) that the CLIPSeg version trained on PC+ performs better than the CLIP-Deconv baseline and the version trained on LVIS, which contains only object labels instead of complex phrases. This result suggests that both dataset variability and model complexity are necessary for generalization. 
ViTSeg performs worse, which is expected as it misses the strong CLIP backbone, known for its generalization capabilities.

\begin{table}
    \centering
    \footnotesize
    \begin{tabular}{llllllll}
        \toprule
         & \multicolumn{2}{c}{Affordances} & \multicolumn{2}{c}{Attributes} & \multicolumn{2}{c}{Meronymy}  \\
        &  $\miou$ &  AP & $\miou$ & AP & $\miou$  & AP \\
        \midrule
CLIPSeg (PC+) & 36.9 & \textbf{50.5} & 26.6 & \textbf{43.0} & \textbf{25.7} & \textbf{29.0} \\ 
CLIPSeg (LVIS)  & \textbf{37.7} & 44.6 & 18.4 & 16.6 & 18.9 & 13.8 \\ 
CLIP-Deconv & 32.2 & 43.7 & 23.1 & 35.6 & 21.1 & 27.1 \\ 
VITSeg (PC+) & 19.2 & 23.5 & \textbf{26.8} & 28.0 & 18.4 & 15.9 \\
         \bottomrule
    \end{tabular}%
    \caption{Performance for generalized prompts. While the PC+-model has seen prompts during training (colliding prompts with test set were removed), the LVIS version was trained on object classes only and is able to generalize due to the CLIP backbone. We use the best threshold $t$ for each model.}
    \label{tab:generalize}
\end{table}

\subsection{Ablation Study}

In order to identify crucial factors for the performance of CLIPSeg, we conduct an ablation study on PhraseCut (Tab.~\ref{tab:ablation}). We evaluate text-based and visual prompt-based performance (obtained using our modifications on PhraseCut) separately for a complete picture.
Both text-based and visual performance drops when random weights instead of CLIP weights are used (``no CLIP pre-training''). When the number of parameters is reduced to 16 (``$D=16$'') performance decreases substantially, which indicates the importance of the information processing in the decoder.
Using an unfavourable visual prompting technique (``highlight mask'') degrades performance on visual input, which supports our findings from Sec.~\ref{sec:prompt_engineering}. 
Using only early activations from layer 3 decreases performance (``only layer 3''), from which we conclude that higher level features of CLIP are useful for segmentation.
Training without visual samples (``no visual'') decreases the performance on visual samples, which is expected as visual and text vectors do not align perfectly. The gap in text-based performance to the hybrid version (PC+) is negligible.

\begin{table}[t]
    \centering
    \footnotesize
    \begin{tabular}{llllllll}
        \toprule
        & \multicolumn{2}{c}{Text-based} & \multicolumn{2}{c}{Visual-based} \\
         & $\miou$ &  AP & $\miou$ &  AP   \\
        \midrule
              CLIPSeg (PC+) &  43.6 &  76.7 &  \textbf{25.4} &  \textbf{55.6} \\
 no CLIP pre-training &  13.1 &  12.6 &  12.7 &   - \\
            no visual &  \textbf{46.4} &  \textbf{77.8} &  14.4 &  31.0 \\
                     $D=16$ &  37.4 &  71.5 &  24.7 &  51.2 \\
         only layer 3 &  31.9 &  64.9 &  21.5 &  48.6 \\
       highlight mask &  43.4 &  75.4 &  23.3 &  43.8 \\
        
        \bottomrule
    \end{tabular}
    \caption{Ablation study conducted on PhraseCut, involving text (left) and visual prompts (right) at test time. We use the best threshold $t$ for each model.}
    \label{tab:ablation}
\end{table}

\section{Conclusion}

We presented the CLIPSeg image segmentation approach that can be adapted to new tasks by text or image prompts at inference time instead of expensive training on new data. 
Specifically, we investigated the novel visual prompt engineering in detail and demonstrated competitive performance on referring expression, zero-shot and one-shot image segmentation tasks.
Beyond that, we showed -- both qualitatively and quantitatively -- that our model generalizes to novel prompts involving affordances and properties.
We expect our method to be useful, especially for inexperienced users for building a segmentation model by specifying prompts and in robotic setups when interaction with humans is desired.
We believe that tackling multiple tasks is a promising direction for future research toward more generic and real-world compatible vision systems.
In a wider context, our experiments, in particular the comparison to the ImageNet-based ViTSeg baseline, highlight the power of foundation models like CLIP for solving several tasks at once.

\paragraph{Limitations}
Our experiments are limited to only a small number of benchmarks, in future work more modalities such as sound and touch could be incorporated. We depend on a large-scale dataset (CLIP) for pre-training. Note, we do not use the best-performing CLIP model ViT-L/14@336px due to weight availability. Furthermore, our model focuses on images, an application to video might suffer from missing temporal consistency. Image size may vary but only within certain limits (for details see supplementary).

\paragraph{Broader Impact}
There is a chance that the model replicates dataset biases from PhraseCut but especially from the unpublished CLIP training dataset. Provided models should be used carefully and not in tasks depicting humans.
Our approach enables adaptation to new tasks without energy-intensive training.

{
\small
\bibliography{literature_clean}
\bibliographystyle{unsrtnat}
}

\clearpage

\section*{Appendix}

\input{supplementary_content}

\end{document}

%% file: supplementary_content.tex
\subsection*{Experimental Setup}

\label{sec:train_setup}

Throughout our experiments we use PyTorch \cite{pytorch} with CLIP ViT-B/16 \cite{radford20}. We train on PhraseCut \cite{wu20phrasecut} for 20,000 iterations on batches of size 64 with an initial learning rate of 0.001 (for VitSeg 0.0001) which decays following a cosine learning rate schedule to 0.0001 (without warmup). We use automatic mixed precision and binary cross entropy as the only loss function.

\subsection*{Image-size Dependency of CLIP}
\begin{figure}[h]
    \centering
    \includegraphics[width=7cm]{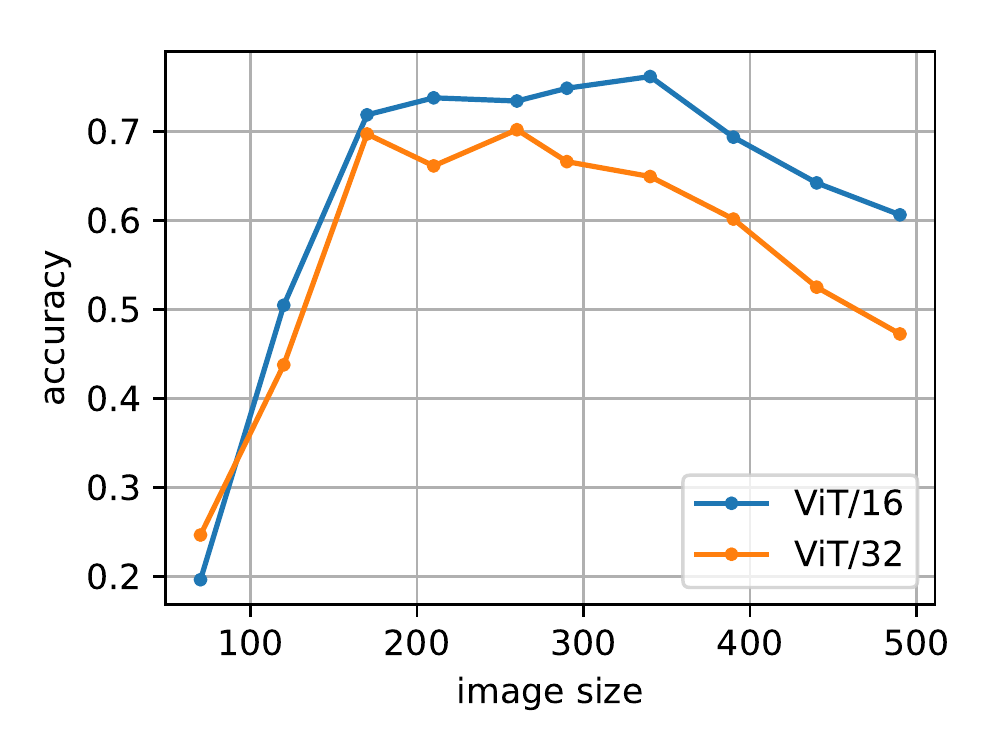}
    \caption{Image classification performance of CLIP over different image sizes.}
    \label{fig:image_size}
\end{figure}

Since multi-head attention does not require a fixed number of tokens, the visual transformer of CLIP can handle inputs of arbitrary size. However, the publicly available CLIP models (ViT-B/16 and ViT-B/32) were trained on 224 $\times$ 224 pixel images. In this experiment we investigate how CLIP performance relates to the input image size -- measured in a classification task.
To this end, we extract the CLS token vector in the last layer from both CLIP models. Using this feature vector as an input, we train a logistic regression classifier on a subset of ImageNet \cite{imagenet} classes differentiating 67 classes of vehicles (Fig. \ref{fig:image_size}).
Our results indicate that CLIP generally handles large image sizes well, with the 16-px-patch version (ViT-B/16) showing a slightly better performance at an optimal image size of around 350 $\times$ 350 pixels.

\subsection*{Object-mapping for Affordances and Attributes}
\label{sec:object_mapping}

For our systematic analysis on generalization (Section 5.5 in the main paper), we generate samples by replacing the following object categories by affordances (bold). 
\\

\noindent Affordances:\\
\textbf{{sit on}}: armchair, sofa, loveseat, deck chair, rocking chair, highchair, deck chair, folding chair, chair, recliner, wheelchair\\
\textbf{{drink from}}: bottle, beer bottle, water bottle, wine bottle, thermos bottle\\
\textbf{{ride on}}: horse, pony, motorcycle\\

\noindent Attributes:\\
\textbf{{can fly}}: eagle, jet plane, airplane, fighter jet, bird, duck, gull, owl, seabird, pigeon, goose, parakeet\\
\textbf{{can be driven}}: minivan, bus (vehicle), cab (taxi), jeep, ambulance, car (automobile)\\
\textbf{{can swim}}: duck, duckling, water scooter, penguin, boat, kayak, canoe\\

\noindent Meronymy (part-of relations):\\
\textbf{{has wheels}}: dirt bike, car (automobile), wheelchair, motorcycle, bicycle, cab (taxi), minivan, bus (vehicle), cab (taxi), jeep, ambulance\\
\textbf{{has legs}}: armchair, sofa, loveseat, deck chair, rocking chair, highchair, deck chair, folding chair, chair, recliner, wheelchair, horse, pony, eagle, bird, duck, gull, owl, seabird, pigeon, goose, parakeet, dog, cat, flamingo, penguin, cow, puppy, sheep, black sheep, ostrich, ram (animal), chicken (animal), person\\

\subsection*{Average Precision Computation}

The average precision metric has the advantage of not depending on a fixed threshold. This is particularly useful when new classes occur which lead to uncalibrated predictions. 
Instead of operating on bounding boxes as in detection, we compute average precision at the pixel-level. This makes the computation challenging, since AP is normally computed by sorting all predictions (hence all pixels) according their likelihood, which requires keeping them in the working memory. For pixels, this is not possible.
To circumvent this, we define a fixed set of thresholds and aggregate statistics (true-positives, etc.) in each image. Finally, we sum up the statistics per threshold level and compute the precision-recall curve. Average precision, which is the area under the precision-recall curve is computed using Simpson integration.



\subsection*{Qualitative Predictions}

\begin{figure*}
    \centering
    \includegraphics[width=0.95\textwidth]{qualitative_clipseg_phrasecut3.pdf}
    \includegraphics[width=0.95\textwidth]{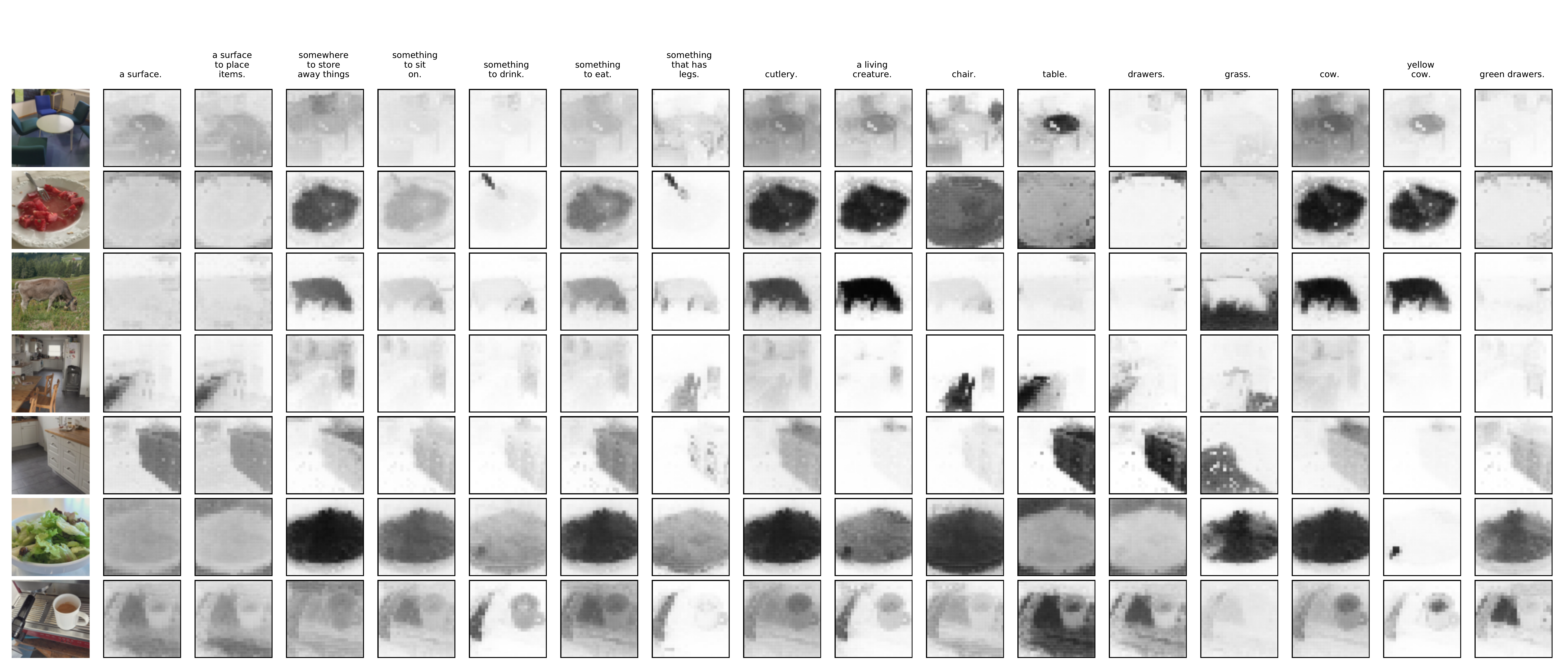}
    \caption{Qualitative predictions of CLIPSeg (PC+) (top, same as Fig.~4 of main paper for reference) and ViTSeg (PC) (bottom).}
    \label{fig:vit_qualitative}
\end{figure*}

In Fig.~\ref{fig:vit_qualitative} we show predictions of ViTSeg (PC), analogous to Fig.~4 of the main paper. In fact, ViTSeg trained with visual samples (PC+) shows worse performance. 
The predictions clearly indicate the deficits of an ImageNet-trained ViT backbone compared to CLIP: Details in the prompt are not reflected by the segmentation and a large number of false positives occur. 

\subsection*{Text prompts, object sizes and classes}

\begin{figure}
    \centering
    \includegraphics[height=3.5cm]{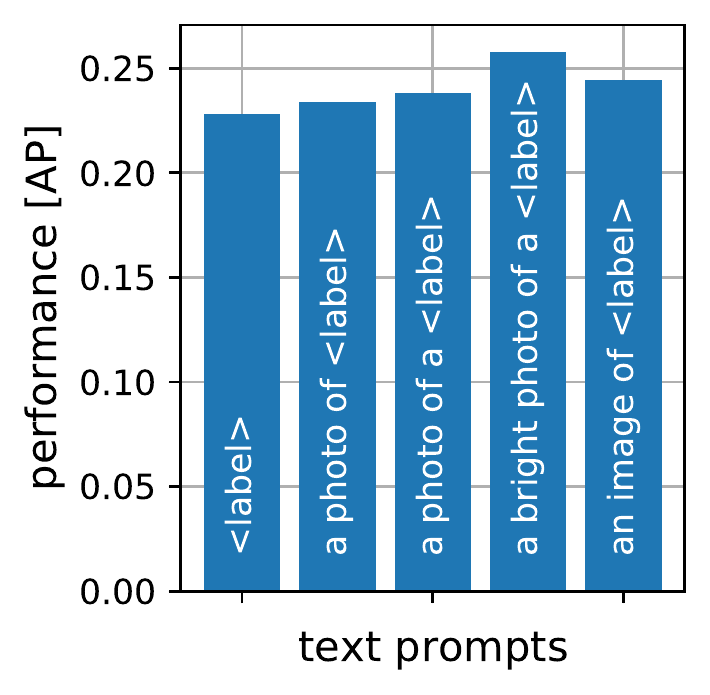}
    \caption{Effect of different text prompts on performance.}
    \label{fig:text_prompts}
\end{figure}

\begin{figure}
    \centering
    \includegraphics[height=3.5cm]{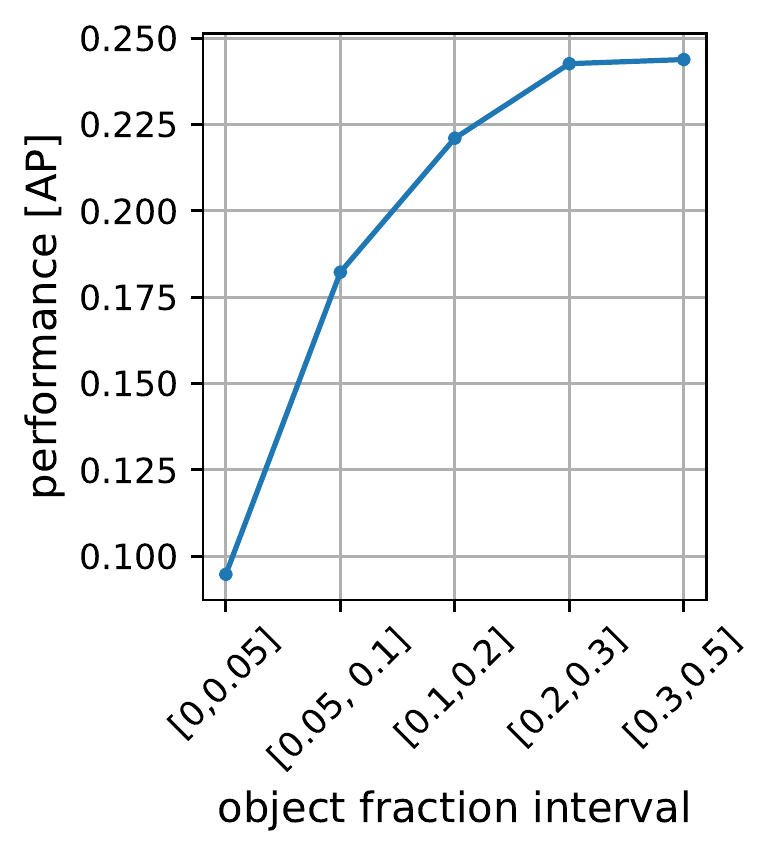}
    \includegraphics[height=3.5cm]{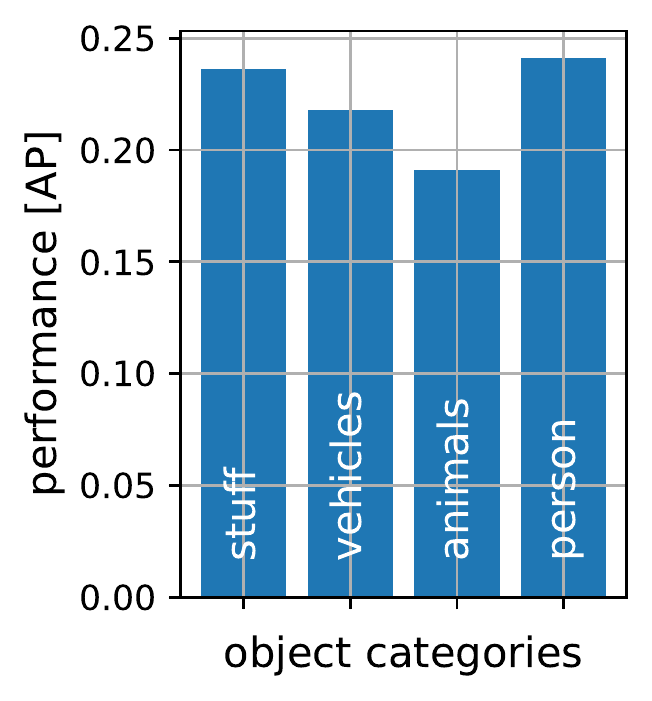}
    \caption{Effect of object size and class on performance.}
    \label{fig:performance_analysis}
    
\end{figure}

To develop a better understanding of when our model performs well, we compare different text prompts (Fig.~\ref{fig:text_prompts}), object sizes (Fig.~\ref{fig:performance_analysis}, left) and object classes (Fig.~\ref{fig:performance_analysis}, right). This evaluation is conducted on a pre-trained  CLIPSeg (PC+). In all cases we randomly sample different prompt forms during training. 
Here we assess the performance on 5,000 samples of the PhraseCut test set.

We see a small effect on performance for alternative prompt forms.
In terms of object size there is a clear trend towards better performance on larger objects.
Performance over different classes is fairly balanced.

